\documentclass{article}
\usepackage{arxiv}
\usepackage[utf8]{inputenc} 
\usepackage[T1]{fontenc}    
\usepackage{hyperref}       
\usepackage{url}            
\usepackage{booktabs}       
\usepackage{amsfonts}       
\usepackage{nicefrac}       
\usepackage{microtype}      
\usepackage{lipsum}		
\usepackage{graphicx}
\usepackage{natbib}
\usepackage{doi}
\usepackage{amsmath}        

\title{Contrastive Bi-Encoder Models for Multi-Label Skill Extraction: Enhancing ESCO Ontology Matching with BERT and Attention Mechanisms}

\author{ \href{https://orcid.org/0000-0000-0000-0000}{\includegraphics[scale=0.06]{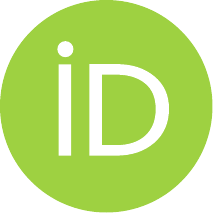}\hspace{1mm}Yongming Sun}\\
	School of Economics\\
	Zhejiang University\\
	Hangzhou, China \\
	\texttt{yongming.sun@zju.edu.cn} \\
}

\renewcommand{\headeright}{}
\renewcommand{\undertitle}{}
\renewcommand{\shorttitle}{}

\begin{document}
\maketitle

\begin{abstract}
Fine-grained labor market analysis increasingly relies on mapping unstructured job advertisements to standardized skill taxonomies such as ESCO. This mapping is naturally formulated as an Extreme Multi-Label Classification (XMLC) problem, but supervised solutions are constrained by the scarcity and cost of large-scale, taxonomy-aligned annotations—especially in non-English settings where job-ad language diverges substantially from formal skill definitions. We propose a zero-shot skill extraction framework that eliminates the need for manually labeled job-ad training data. The framework uses a Large Language Model (LLM) to synthesize training instances from ESCO definitions, and introduces hierarchically constrained multi-skill generation based on ESCO Level-2 categories to improve semantic coherence in multi-label contexts. On top of the synthetic corpus, we train a contrastive bi-encoder that aligns job-ad sentences with ESCO skill descriptions in a shared embedding space; the encoder augments a BERT backbone with BiLSTM and attention pooling to better model long, information-dense requirement statements. An upstream RoBERTa-based binary filter removes non-skill sentences to improve end-to-end precision. Experiments show that (i) hierarchy-conditioned generation improves both fluency and discriminability relative to unconstrained pairing, and (ii) the resulting multi-label model transfers effectively to real-world Chinese job advertisements, achieving strong zero-shot retrieval performance (F1@5 = 0.72) and outperforming TF--IDF and standard BERT baselines. Overall, the proposed pipeline provides a scalable, data-efficient pathway for automated skill coding in labor economics and workforce analytics.
\end{abstract}


\section{Introduction}

Understanding labor market dynamics increasingly requires measurement at the skill level \citep{turrell2019nberw25837}. Standardized taxonomies such as the European Skills, Competences, Qualifications and Occupations (ESCO) \citep{ec_esco_website} provide a controlled vocabulary that supports comparability across firms, regions, and time. However, real-world job advertisements are written in informal, heterogeneous language and often express requirements implicitly (e.g., through tasks, tools, or domain context). Bridging this gap—mapping job-ad text to a large set of standardized skills—naturally yields an Extreme Multi-Label Classification (XMLC) problem with a very large label space and strong label correlations.

The main obstacle for supervised XMLC in this domain is data availability. Training strong discriminative models typically requires large corpora of job sentences annotated with taxonomy-specific skill codes, which are expensive and slow to construct, difficult to maintain as taxonomies evolve, and especially scarce for non-English labor markets. As a result, many practical systems fall back on keyword rules or generic similarity matching. These approaches are brittle under vocabulary mismatch (synonyms, paraphrases, and local jargon) and struggle to capture compositional and contextual skill expressions (e.g., ``experience deploying models in production'' implying MLOps-related skills).

This paper proposes a \textbf{zero-shot} pipeline that bypasses manual job-ad annotation while still learning a skill-aware representation aligned to ESCO. The core idea is to use ESCO skill definitions as supervision: we synthesize training instances with a Large Language Model (LLM) and train a contrastive bi-encoder to align (sentence, skill-definition) pairs in a shared embedding space. A key design choice is \textbf{hierarchically constrained generation}: instead of sampling arbitrary multi-skill combinations, we condition multi-skill synthesis on ESCO Level-2 categories to induce more realistic co-occurrence structure and reduce semantic drift. In addition, we include an upstream binary filter to discard non-skill sentences before retrieval, improving precision and computational efficiency in end-to-end processing.

Our main contributions are:
\begin{itemize}
    \item \textbf{Hierarchy-conditioned synthetic supervision:} We propose an LLM-based data generation strategy that leverages ESCO's hierarchy (Level-2 constraints) to produce more semantically coherent multi-skill training instances than unconstrained pairing.
    \item \textbf{Contrastive zero-shot XMLC for skills:} We train a contrastive bi-encoder that aligns job-ad sentences with ESCO skill definitions without using any manually labeled job-ad training data, enabling scalable zero-shot retrieval over large skill sets.
    \item \textbf{An end-to-end extraction pipeline:} We combine a RoBERTa-based binary sentence filter with multi-label retrieval, and show strong transfer to real-world Chinese job advertisements with substantial gains over TF--IDF and standard BERT baselines.
\end{itemize}

The remainder of the paper is organized as follows. Section~\ref{sec:LR} reviews related work. Section~\ref{sec:data} introduce our data. Section~\ref{sec:method} describes the proposed pipeline, including data generation and model training. Section~\ref{sec:experiment} introduces the experimental setup. Section~\ref{sec:result} reports results, ablations, and robustness analyses. Section~\ref{sec:conclusion} concludes.

\section{Related Work} \label{sec:LR}

Our work connects three strands of research: (i) extreme multi-label text classification (XMLC), (ii) skill extraction for labor market intelligence, and (iii) LLM-driven synthetic supervision and contrastive retrieval. Below, we summarize the most relevant lines of work and clarify how our framework departs from existing approaches.

\subsection{Extreme Multi-Label Text Classification (XMLC)}

XMLC concerns prediction over very large label spaces, often with severe label imbalance and strong label correlations. Classical solutions used sparse linear one-vs-rest formulations \citep{babbar2017dismec, yen2016pdsparse}, including methods that emphasized optimization and sparsity to make training feasible at scale. While effective in some regimes, these approaches typically rely on surface-feature engineering and do not explicitly model label semantics, making transfer to new domains or label sets difficult.

More recent work has adopted deep and pretrained language-model (PLM) backbones. A prominent direction uses tree or graph structures over labels to reduce computational cost at inference time \citep{prabhu2018parabel}, while another line directly leverages Transformer encoders with attention mechanisms to improve ranking quality \citep{you2019attentionxml, jiang2021lightxml}. Despite strong supervised performance, most state-of-the-art XMLC systems assume access to large amounts of labeled data and are sensitive to distribution shifts.

In contrast, our setting is explicitly \emph{zero-shot} with respect to job-ad labels: we aim to predict ESCO skills without manually annotated job advertisements. Instead of learning label-specific classifiers from labeled examples, we cast XMLC as a retrieval problem in a shared semantic space, where the label side is represented by ESCO definitions and related textual metadata. This choice naturally supports scalability to large ontologies and mitigates the dependency on task-specific annotation.

\subsection{Skill Extraction and Labor Market Intelligence}

Skill extraction is a core component of labor market intelligence, enabling measurement of changing demand, skill complementarities, and occupational transitions. Earlier pipelines commonly formulated skill detection as named entity recognition (NER) or dictionary matching, sometimes augmented with rule-based normalization. These methods are interpretable and efficient, but they are brittle to paraphrase, domain drift, and the long-tail of job-ad language.

Supervised learning approaches (e.g., sequence labeling or multi-class classification) \citep{zhang2022skillws, le2019skill2vec} improve recall and robustness but are constrained by the availability and coverage of annotated corpora. This bottleneck is particularly acute for comprehensive taxonomies such as ESCO or O*NET, where label sets are large and definitions are fine-grained.

A further practical challenge is \emph{vocabulary mismatch}: job ads frequently use informal phrasing, abbreviations, and firm-specific terminology that diverge from ontology definitions. Several studies address this via similarity-based matching between job text and taxonomy entries \citep{bhola2020retrievingskills, decorte2021jobbert}; however, naïve similarity metrics often fail under paraphrase and do not learn domain-appropriate alignment. Our approach targets this mismatch by learning a contrastive bi-encoder that embeds job-ad sentences and ESCO skill definitions into a unified representation space, explicitly optimizing semantic alignment rather than relying on fixed lexical overlap.

\subsection{LLM-Driven Synthetic Supervision}

The lack of labeled data has motivated synthetic supervision via large language models (LLMs) \citep{ding2023gpt3annotator, wang2021gpt3labelingcost, west2022symbolickd, ye2022zerogen}, including prompt-based augmentation, self-instruction, and automatic labeling. Across many NLP tasks, synthetic data can yield competitive models when real annotation is costly, provided that the synthetic distribution is sufficiently diverse and the labeling rules are reliable.

Within skill extraction, the main difficulty is not only generating fluent text but generating \emph{task-faithful} samples that reflect realistic co-occurrence patterns and decision boundaries. Purely random or unconstrained generation can lead to artifacts (e.g., overly stereotyped phrasing, unrealistic label combinations, or overly clean separability), which in turn produces models that appear strong on synthetic validation but transfer poorly to real job ads.

We address this by introducing a \emph{hierarchically constrained} generation strategy that conditions the LLM on ESCO Level-2 structure. Constraining multi-skill samples to remain within coherent ESCO subtrees encourages more plausible co-occurrence and reduces spurious correlations. In effect, the ESCO hierarchy acts as distant supervision: it injects structural priors into synthetic data so that the learned model better reflects the ontology’s semantics while remaining agnostic to any manually labeled job-ad corpus.

\subsection{Contrastive Learning for Semantic Retrieval}

Contrastive learning has become a standard approach for training sentence and passage embeddings for semantic retrieval \citep{gao2021simcse}. Bi-encoder architectures \citep{reimers2019sentencebert} are particularly attractive in large-label settings because they support efficient approximate nearest-neighbor search and decouple encoding of queries (job sentences) and candidates (skill definitions). Typical training objectives pull matched pairs together while pushing apart negatives sampled within a batch or from hard-negative mining \citep{robinson2021hcl}.

We adopt this retrieval framing for XMLC (sentence-to-skill matching) and implement a contrastive objective that treats a sentence and its target ESCO skill definition as a positive pair. Importantly, negative sampling is central in this setting: the model must discriminate between semantically adjacent skills (hard negatives), not only between unrelated concepts. Our ablation results highlight how margin choice and the number of negatives govern the trade-off between ranking quality and training efficiency.

Finally, while many bi-encoder systems rely solely on the \texttt{[CLS]} representation, job-ad sentences often contain long, compositional requirements with multiple constraints. We therefore augment the PLM backbone with lightweight sequence modeling and pooling (e.g., BiLSTM plus attention pooling) \citep{conneau2017nliuniversal} to better aggregate distributed evidence across the sentence, improving robustness when skill cues are not localized to a single token span.

\section{Data} \label{sec:data}

Our study utilizes two primary data sources: a standardized skills taxonomy for synthetic generation and a large-scale corpus of real-world job advertisements for evaluation and application.

\subsection{Standardized Taxonomy: ESCO}

We adopt the \textit{European Skills, Competences, Qualifications and Occupations} (ESCO) ontology (v1.1) as our ground truth for skill definitions. ESCO provides a comprehensive, hierarchical classification of skills relevant to the EU labor market but applicable globally. The taxonomy is structured into four levels, ranging from broad categories (e.g., "Language and communication") to granular skills.

For this study, we focus on the leaf nodes of the hierarchy, specifically the \textbf{Level-4 skills}, resulting in a total of \textbf{13,890 distinct skill concepts}. Each concept is associated with a preferred label and a textual description (e.g., \textit{"Python (computer programming): The techniques and principles of software development, such as analysis, algorithms, coding, testing and compiling of programming paradigms in Python."}). These descriptions serve as the seed text for our synthetic data generation pipeline.

\subsection{Real-World Job Advertisements}

To evaluate the zero-shot transfer capabilities of our model, we curate a dataset of real-world job postings collected from Zhaopin.com, one of the largest online recruitment platforms in China. The raw corpus spans a nine-year period from \textbf{2015 to 2023}, covering a diverse range of industries, regions, and job types.

From this massive collection, we construct a representative experimental dataset by randomly sampling \textbf{200,000 job advertisements}. Each advertisement contains unstructured text fields describing job responsibilities and requirements. We pre-process this text by segmenting it into individual sentences based on punctuation (periods, question marks, and exclamation points), resulting in a pool of variable-length sentences that mix skill requirements with general administrative information (e.g., "salary negotiable", "location: Beijing"). This real-world dataset presents significant challenges, including colloquialisms, implied skills, and noisy formatting, which our pipeline is designed to address.

\section{Methodology} \label{sec:method}

We propose an Extreme Multi-Label Classification (XMLC) framework that maps real-world job vacancy text to the standardized European Skills, Competences, Qualifications and Occupations (ESCO) skill ontology. Because large-scale human-annotated job posting data are scarce, our approach is \emph{annotation-free on real postings}: we rely on synthetic supervision constructed from ESCO and an LLM, and then transfer the learned semantic matching capability to real job ads.

Given a job ad, we segment it into sentences and perform a three-stage workflow: (1) synthetic data generation and pre-processing, (2) contrastive bi-encoder training, and (3) inference via semantic retrieval and aggregation.

\subsection{Synthetic Data Generation}

\paragraph{ESCO localization.}
ESCO (v1.1.2) provides 13,896 distinct skills. Since our target corpus consists of Chinese job postings, we manually translate ESCO skill names and definitions into Chinese and use the translated descriptions consistently throughout training and inference.

\paragraph{Positive samples.}
We construct synthetic sentences using the DeepSeek-V3 LLM conditioned on ESCO skills. We generate two types of positive samples to reflect different supervision granularity:
\begin{itemize}

    \item \textbf{Single-skill samples ($D_{\text{single}}$).} For each skill $s \in \mathcal{S}$, we generate a set of job-ad-like sentences $T_s=\{t_1,t_2,\dots\}$ that express skill $s$ in realistic requirement language. The resulting training pairs are $(t, s)$.
    
    \item \textbf{Co-occurring skill samples ($D_{\text{multi}}$).} To model realistic multi-skill requirements, we generate sentences conditioned on skill pairs $(s_i,s_j)$. To encourage semantic coherence, we sample pairs from the same Level-2 ESCO category as a weak structural constraint. The resulting samples are $(t, \{s_i,s_j\})$.
    
\end{itemize}

We describe the per-skill generation count $|T_s|$ later (and provide prompt templates in the appendix).

\paragraph{Negative samples ($D_{\text{none}}$).}
To train a sentence-level filter that removes non-requirement content, we additionally generate negative sentences typical of job ads (e.g., company introductions, benefits, compliance statements) that contain no explicit skill requirements.

\paragraph{Decoding settings and quality control.}
We use stochastic decoding with temperature $T=0.7$ and nucleus sampling top-$p=0.9$ (maximum 128 tokens per sentence) to encourage lexical diversity while preserving instruction-following. To mitigate near-duplicates and style artifacts:
\begin{itemize}
    \item \textbf{De-duplication.} We remove exact duplicates and near-duplicates by thresholding cosine similarity between sentence embeddings (Sentence-BERT) and discarding samples above a high similarity cutoff.
    \item \textbf{Diversity enforcement.} For each skill, we enforce minimum lexical variation (e.g., avoiding repeated n-grams across generated sentences) and resample when violations occur.
\end{itemize}

\paragraph{Handling generic or overlapping skills.}
ESCO includes skills whose definitions are broad or lexically similar (e.g., generic soft skills). Such skills can induce ambiguous supervision and unstable retrieval. We therefore apply two safeguards:
\begin{itemize}
    \item \textbf{Ambiguity-aware sampling.} For skills whose translated definitions have unusually high similarity to many other skills, we increase diversity constraints and require generated sentences to include job-context anchors (role/task cues) beyond generic phrasing.
    \item \textbf{Evaluation-time disambiguation.} During retrieval, we rely on similarity thresholding (Section~\ref{sec:inference}) and sentence-to-posting aggregation to reduce spurious activation from generic sentences.
\end{itemize}

\subsection{Data Pre-processing and Filtering}
\label{sec:filter}

Real-world job postings contain substantial non-requirement content. Before skill retrieval, we apply a sentence-level binary filter to retain only sentences that are likely to express skill requirements.

We fine-tune a RoBERTa-base classifier on a balanced dataset with $D_{\text{single}}$ as positives and $D_{\text{none}}$ as negatives, using standard binary cross-entropy. During inference, a sentence $t$ is discarded if

\begin{equation}
    P(\text{skill}\mid t) < \tau,
\end{equation}

where $\tau$ is selected on a held-out validation set to achieve a desired precision--recall trade-off (we prioritize high precision to reduce downstream false positive skills).

\subsection{Contrastive Bi-Encoder Architecture}

We formulate XMLC as semantic retrieval: embed a job-ad sentence and each candidate ESCO skill into a shared vector space, then retrieve the nearest skills.

\subsubsection{Encoder Backbone}

We use a Siamese bi-encoder with \emph{shared parameters} for both towers. The encoder $\phi(\cdot)$ is initialized with \texttt{bert-base-chinese}. Given an input sequence $x$ (either a job-ad sentence $t$ or a translated skill description $s$), we extract contextualized token embeddings $H \in \mathbb{R}^{L \times h}$ from the last hidden layer, where $L$ is the sequence length and $h=768$ is the hidden size. To better capture local sequential patterns and salient keywords in skill-related text, we augment the Transformer outputs with a BiLSTM and an attention pooling layer \citep{yang2016han}:

\begin{enumerate}
\item \textbf{BiLSTM.} We pass $H$ through a bidirectional LSTM:
\begin{equation}
    H_{\mathrm{lstm}} = \mathrm{BiLSTM}(H) \in \mathbb{R}^{L \times 2h'},
\end{equation}
where $h'=256$ is the hidden size of each LSTM direction.

\item \textbf{Attention pooling.} We compute a weighted aggregation of the LSTM outputs to form a fixed-size representation. Concretely, we first compute a scalar score for each token position $\ell \in \{1,\dots,L\}$:
\begin{equation}
    u_\ell = w^\top \tanh\!\left(W H_{\mathrm{lstm},\ell}\right),
\end{equation}
where $H_{\mathrm{lstm},\ell} \in \mathbb{R}^{2h'}$ denotes the $\ell$-th token vector, $W \in \mathbb{R}^{a \times 2h'}$ and $w \in \mathbb{R}^{a}$ are trainable parameters, and $a$ is the attention hidden dimension. We then normalize scores with a softmax over token positions:
\begin{equation}
    \alpha_\ell = \frac{\exp(u_\ell)}{\sum_{j=1}^{L} \exp(u_j)}.
\end{equation}
The pooled representation is
\begin{equation}
    v = \sum_{\ell=1}^{L} \alpha_\ell H_{\mathrm{lstm},\ell} \in \mathbb{R}^{2h'}.
\end{equation}

\item \textbf{Projection.} The final embedding is obtained via a linear projection followed by $L_2$ normalization:
\begin{equation}
    e(x)=\frac{W_p v + b}{\|W_p v + b\|_2} \in \mathbb{R}^{d},
\end{equation}
where $W_p \in \mathbb{R}^{d \times 2h'}$, $b \in \mathbb{R}^{d}$, and we set $d=128$. We denote sentence embeddings as $e_t=e(t)$ and skill embeddings as $e_s=e(s)$.
\end{enumerate}

\subsection{Training Objective}

We train the model with a margin-based contrastive ranking objective. For a positive pair $(t, s^+)$, we sample $K$ negative skills \citep{robinson2021hcl} $\{s^-_1,\dots,s^-_K\}$ such that $s^-_k \notin \mathcal{Y}(t)$, where $\mathcal{Y}(t)$ is the set of ground-truth skills used to generate $t$. Let $\mathrm{sim}(\cdot,\cdot)$ denote cosine similarity. The loss is
\begin{equation}
\mathcal{L}(t,s^+) = \frac{1}{K}\sum_{k=1}^{K} \max\Bigl(0, \lambda - \mathrm{sim}(e_t,e_{s^+}) + \mathrm{sim}(e_t,e_{s^-_k})\Bigr),
\end{equation}
where $\lambda=0.5$ is the margin. For multi-label samples $(t,\{s_1,s_2\})$, we average over positives:
\begin{equation}
    \mathcal{L}_{\text{multi}}(t,\{s_1,s_2\})=\frac{1}{2}\bigl(\mathcal{L}(t,s_1)+\mathcal{L}(t,s_2)\bigr).
\end{equation}

\subsection{Inference and Retrieval}
\label{sec:inference}

We pre-compute embeddings for all translated ESCO skills $\mathcal{S}$, yielding an index matrix $E_{\mathcal{S}}\in\mathbb{R}^{|\mathcal{S}|\times d}$. For each filtered sentence $t$, we compute $e_t$ and retrieve candidates by cosine similarity. We then apply a similarity threshold to control precision:
\begin{equation}
\hat{\mathcal{S}}(t)=\Bigl\{\, s \in \mathrm{TopK}_{s'\in\mathcal{S}} \ \mathrm{sim}(e_t,e_{s'}) \ \Bigm|\ \mathrm{sim}(e_t,e_s)\ge \gamma \Bigr\},
\end{equation}
where $K$ is a fixed retrieval budget and $\gamma$ is tuned on validation data.

\paragraph{Posting-level aggregation.}
Given a job posting consisting of sentences $\{t_m\}_{m=1}^{M}$, we produce the posting-level predicted skill set via union aggregation:
\begin{equation}
    \hat{\mathcal{S}}_{\text{post}}=\bigcup_{m=1}^{M} \hat{\mathcal{S}}(t_m)
    \label{eq:union}
\end{equation}
This union rule reflects the fact that job postings typically list multiple independent requirements across different sentences.

\section{Experiments} \label{sec:experiment}

We conduct a comprehensive series of experiments to evaluate our \emph{annotation-free transfer} pipeline for mapping real-world Chinese job vacancies to standardized ESCO skills. The experiments are designed to answer three questions: (i) whether the LLM-generated synthetic supervision is fluent and diverse, (ii) whether the sentence-level filter effectively removes non-requirement content without suppressing genuine requirements, and (iii) whether the contrastive bi-encoder learns a robust alignment between job-ad language and ESCO skill definitions that transfers to real postings.

\subsection{Experimental Setup}

All experiments are conducted on a single NVIDIA RTX 4090 GPU using PyTorch 2.0. Unless otherwise stated, we use a learning rate of $2 \times 10^{-5}$, batch size 32, and train for 10 epochs. The embedding dimension is fixed to $d=128$. In contrastive training, each positive pair is trained against $K=5$ sampled negative skills. We set the random seed to 42 for initialization and data splitting.

We compare three models:

\begin{itemize}
    \item \textbf{Model A}: Standard BERT (Single Label) -- A baseline using BERT for single-skill sentence embedding with contrastive loss.
    \item \textbf{Model B}: BERT + BiLSTM + Attention (Single Label) -- An optimized single-label model incorporating BiLSTM for sequence modeling and attention for contextual weighting.
    \item \textbf{Model C}: BERT + BiLSTM + Attention (Multi Label) -- Extends Model B to multi-label training, handling one sentence mapped to multiple skills.
\end{itemize}

Across all experiments involving real job ads, evaluation is performed at the \emph{posting level}. We segment each posting into sentences, apply the binary filter with threshold $\tau$, retrieve candidate skills for each retained sentence, and then aggregate to a posting-level prediction by set union in Eq. \ref{eq:union}. Retrieval uses a two-stage decision rule consistent with our Methodology: we first retrieve Top-$K_r$ candidates by cosine similarity, and then keep only those with similarity above $\gamma$. Unless stated otherwise, $K_r=50$ is used as the retrieval budget, while $\gamma$ is tuned on a development split (see Experiment~5).

\subsection{Experiment 1: Synthetic Data Quality and Diversity}

We first assess whether synthetic sentences resemble real job-ad requirement language in both fluency and diversity. We compare three synthetic variants: single-skill sentences ($D_{\text{single}}$), hierarchically constrained multi-skill sentences ($D_{\text{multi}}$), and randomly paired multi-skill sentences (Random-$D_{\text{multi}}$).

For fluency, we compute perplexity on 1{,}000 randomly sampled synthetic sentences using a Chinese GPT-2 language model (\texttt{uer/gpt2-chinese-cluecorpussmall}); lower perplexity indicates more natural phrasing. Since fluency alone does not guarantee usefulness as supervision, we also evaluate transferability: for each variant, we train the same RoBERTa sentence classifier (skill vs.\ non-skill) using that variant as the positive class and a common negative set, and then test on a manually labeled set of 1{,}000 real Chinese sentences drawn from job ads. We report perplexity and real-data F1, where higher F1 suggests the synthetic variant better matches real requirement semantics. To ensure that improvements are not driven by duplicated phrasing, we additionally report the fraction of near-duplicate samples removed by our de-duplication procedure.

\subsection{Experiment 2: Binary Classification for Filtering}

This experiment evaluates the sentence-level filter used to remove non-requirement content (e.g., company introductions, benefits, compliance clauses) prior to skill retrieval. We fine-tune RoBERTa classifiers under three training mixes: positives from $D_{\text{single}}$, positives from $D_{\text{multi}}$, and the combined positives ($D_{\text{single}} \cup D_{\text{multi}}$), each paired with the same negative set $D_{\text{none}}$ (100{,}000 synthetic non-skill sentences). We evaluate on 1{,}000 manually annotated real Chinese sentences segmented by punctuation.

Threshold selection is aligned with the overall pipeline objective: we choose $\tau$ on a held-out development set to \emph{maximize precision subject to a minimum recall constraint}. This reflects the practical need to reduce downstream false positive skills while retaining most true requirement sentences. We compare against two baselines: a keyword-matching heuristic and a zero-shot NLI classifier (Chinese). Results are reported using Accuracy, Precision, Recall, F1, and AUPRC. Finally, we include an ablation that removes $D_{\text{none}}$ during training to quantify the role of explicit non-skill supervision in controlling false positives.

\subsection{Experiment 3: Retrieval Performance on Synthetic Hold-out}

We next evaluate whether the bi-encoder learns a meaningful alignment between synthetic job-ad sentences and ESCO skill definitions. We use a 20\% hold-out split of the synthetic data and evaluate Models A--C as pure retrieval systems. For each query sentence, we rank all ESCO skills by cosine similarity and report standard ranking metrics including MRR, Recall@5, and mAP. This experiment functions as a controlled diagnostic: it tests whether architectural choices (BiLSTM+attention) and supervision choices (single-skill vs.\ multi-skill) improve semantic matching under known ground truth.

To provide a non-neural retrieval reference point, we also report TF-IDF matching between sentences and ESCO definitions, using cosine similarity in TF-IDF space to retrieve Top-$K_r$ candidates.

\subsection{Experiment 4: Component Ablation Study}

To isolate which design choices drive performance, we ablate both architecture and objective components. Starting from Model C, we (i) remove the BiLSTM layer, (ii) replace attention pooling with mean pooling or \texttt{[CLS]} pooling, and (iii) vary contrastive loss hyperparameters by sweeping the margin $\lambda \in \{0.3, 0.5, 0.7\}$ and the number of negative skills $K \in \{1, 5, 10\}$. We evaluate each variant using MRR and Recall@5 on the synthetic hold-out set, and we additionally report training throughput (samples/second) to quantify computational overhead.

\subsection{Experiment 5: End-to-End Transfer to Real Job Ads}

This is the primary evaluation of real-world utility. We test the full pipeline—sentence segmentation, filtering, retrieval with thresholding, and posting-level union aggregation—on 500 randomly sampled real Chinese job advertisements. Each posting is manually annotated with ESCO Level-4 skill IDs based on the translated ESCO ontology. To improve reliability, we employ a two-stage annotation protocol: annotators follow a written guideline to map requirement statements to ESCO IDs, each sample is independently labeled, and disagreements are resolved by adjudication; we report inter-annotator agreement in the appendix.

We tune the retrieval similarity threshold $\gamma$ on a development subset of the annotated postings. Concretely, for each candidate $\gamma$, we run the full pipeline and select the value that maximizes posting-level F1@5 (equivalently, a precision--recall trade-off that best matches the downstream objective). Final results are reported on a disjoint test subset.

We compare against lexical retrieval baselines that match job text to ESCO definitions: TF-IDF cosine similarity (primary baseline) and BM25 (where applicable). We report posting-level Precision@K, Recall@K, and F1@K for $K \in \{1,3,5\}$.

\subsection{Experiment 6: Scalability and Robustness}

Finally, we examine robustness to noisy real-world text and runtime scalability. For robustness, we inject character-level perturbations into the real test set by applying random substitutions and deletions to 20\% of characters in each posting, reflecting common typos and low-quality text. We report the relative degradation in posting-level F1@K and in sentence-level ranking metrics.

For scalability, we measure training convergence and retrieval latency as the synthetic training set size increases from 10k to 100k samples. Retrieval latency is reported as end-to-end time per posting (encoding + similarity search over $|\mathcal{S}|=13{,}896$ skills), measured under the same hardware and batching settings used elsewhere.

\section{Results} \label{sec:result}

This section evaluates the proposed zero-shot skill extraction framework in three steps. We first examine the quality of the synthetic training data, then evaluate the skill-requiring sentence filter, and finally compare the XMLC models (Models A, B, and C) on synthetic benchmarks and real-world job advertisements.

\subsection{Quality of Synthetic Data}

We hypothesize that hierarchically constrained generation produces synthetic sentences that are (i) linguistically fluent and (ii) semantically discriminative for downstream supervision. We evaluate fluency using GPT-2 perplexity and assess discriminability via a binary separability test (ROC curves).

\begin{figure}[h]
	\centering
	\includegraphics[width=\linewidth]{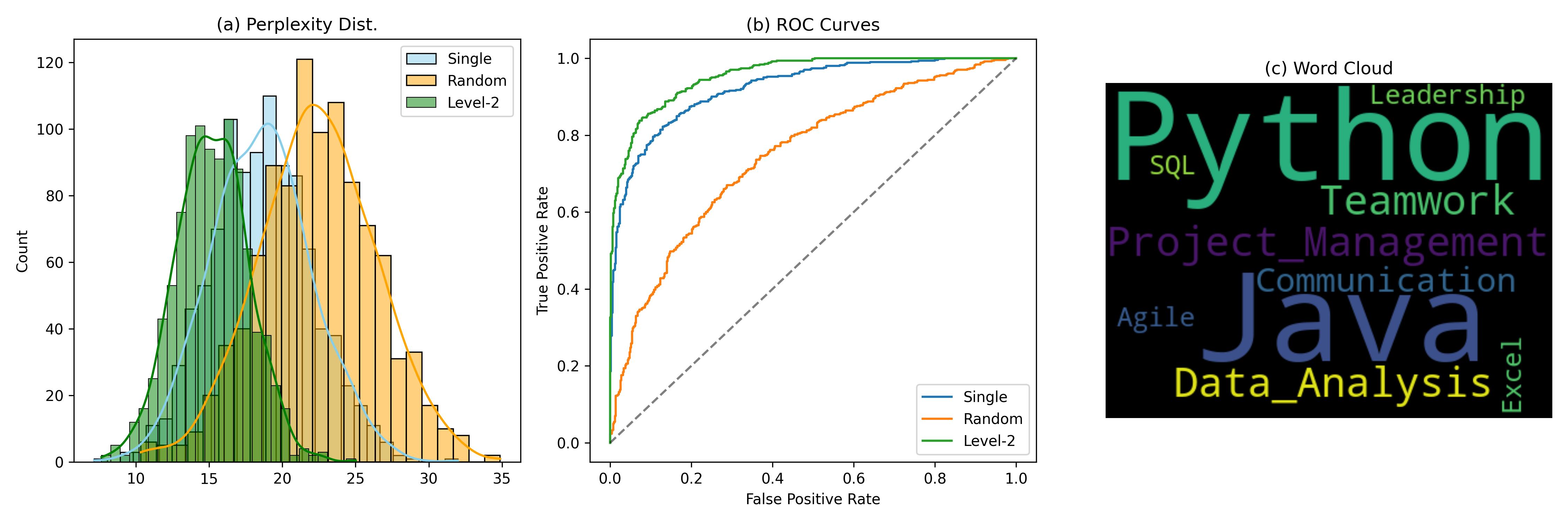}
	\caption{Evaluation of synthetic data quality. (a) GPT-2 perplexity distributions. (b) ROC curves from a separability test between synthetic skill sentences and generic non-skill text; the Level-2 constrained variant dominates across thresholds. (c) Word cloud illustrating broad topical coverage of the generated corpus.}
	\label{fig:synth_fig}
\end{figure}

Table~\ref{tab:synth_quality} reports the perplexity statistics. The \textbf{Multi-Skill (Level-2)} variant achieves the lowest perplexity, indicating that ESCO Level-2 constraints improve linguistic consistency compared to unconstrained multi-skill pairing and single-skill generation. In addition, Figure~\ref{fig:synth_fig}(b) shows that the Level-2 variant yields the strongest ROC curve in the separability test, suggesting that taxonomy-guided generation produces more discriminative positives (i.e., easier to separate from generic non-skill text) and reduces label noise.

\begin{table}[h]
\centering
\caption{Intrinsic quality assessment of synthetic variants using GPT-2 perplexity (lower is better). Discriminability is evaluated separately via ROC curves in Figure~\ref{fig:synth_fig}(b).}
\begin{tabular}{lc}
\hline
\textbf{Synthetic Variant} & \textbf{Perplexity (Mean $\pm$ SD)} \\
\hline
Single-Skill ($D_{single}$) & $18.5 \pm 3.5$ \\
Multi-Skill (Random) & $22.1 \pm 4.0$ \\
\textbf{Multi-Skill (Level-2)} & $\mathbf{15.2 \pm 2.5}$ \\
\hline
\end{tabular}
\label{tab:synth_quality}
\end{table}

\subsection{Effectiveness of Skill-Requiring Sentence Filtering}

\begin{figure}[h]
	\centering
	\includegraphics[width=\linewidth]{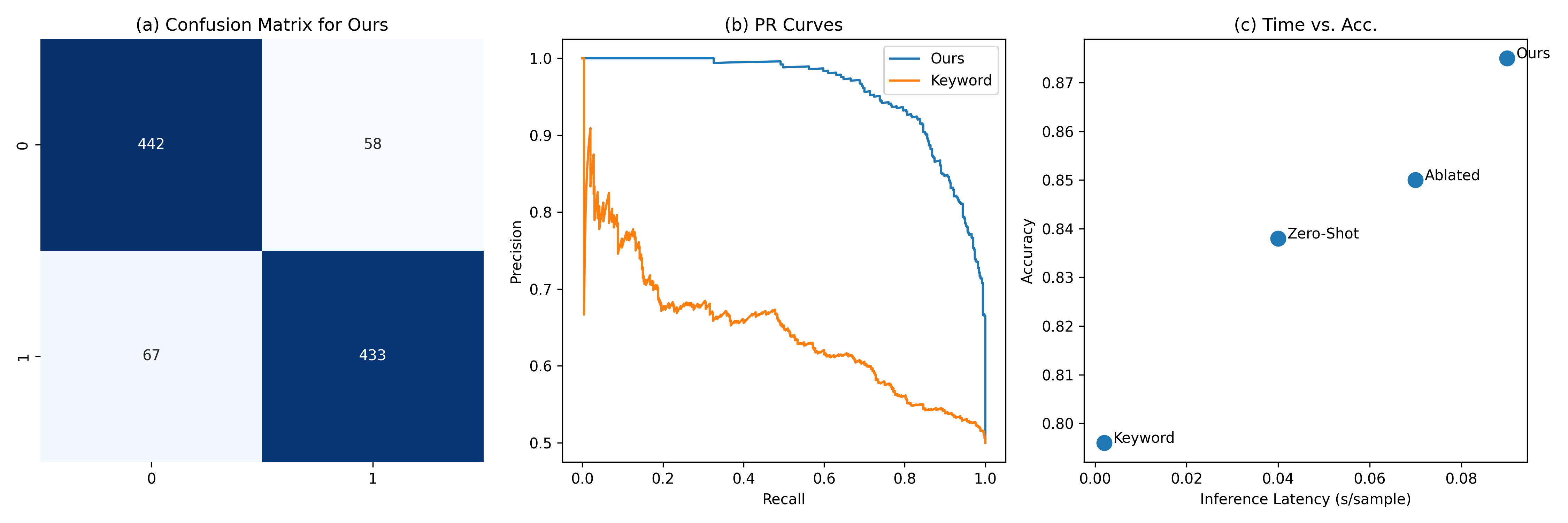}
	\caption{Binary sentence filtering. (a) Confusion matrix for our filter, showing balanced FP/FN counts. (b) Precision--Recall curves comparing our filter to keyword matching across thresholds. (c) Inference latency vs. accuracy for competing approaches.}
	\label{fig:binary_fig}
\end{figure}

A reliable pre-filter is essential because real job ads contain substantial non-skill content (e.g., company introductions and benefits). Figure~\ref{fig:binary_fig} reports the performance of our RoBERTa-based binary classifier. From the confusion matrix in Figure~\ref{fig:binary_fig}(a), our model produces 58 false positives and 67 false negatives (out of 1,000), corresponding to an Accuracy of 0.875, Precision of 0.882, Recall of 0.866, and F1 of 0.874 at the default decision threshold.

Figure~\ref{fig:binary_fig}(b) further shows that our model maintains higher precision over a wide recall range relative to keyword matching. Figure~\ref{fig:binary_fig}(c) summarizes the latency--accuracy trade-off: our model is slower than keyword rules but yields materially higher accuracy, which is appropriate for offline/large-batch processing. Importantly, removing synthetic negative samples ($D_{none}$) increases spurious activations (higher false positives), indicating that explicit ``no-skill'' supervision is necessary to calibrate the filter.

\begin{table}[h]
\centering
\caption{Binary sentence filter performance on real-world Chinese job ads. For \textbf{Ours (Combined)}, Precision/Recall/F1 are computed from the confusion matrix in Figure~\ref{fig:binary_fig}(a).}
\begin{tabular}{lcccc}
\hline
\textbf{Method} & \textbf{Accuracy} & \textbf{F1-Score} & \textbf{Precision} & \textbf{Recall} \\
\hline
Keyword Matching & 0.80 & 0.75 & 0.72 & 0.78 \\
Zero-Shot NLI & 0.84 & 0.80 & 0.79 & 0.81 \\
Ablated (w/o $D_{none}$) & 0.85 & 0.83 & 0.82 & 0.84 \\
\textbf{Ours (Combined)} & \textbf{0.875} & \textbf{0.874} & \textbf{0.882} & \textbf{0.866} \\
\hline
\end{tabular}
\label{tab:binary_perf}
\end{table}

\subsection{XMLC Model Comparison (Synthetic Benchmark)}

\begin{figure}[h]
	\centering
	\includegraphics[width=\linewidth]{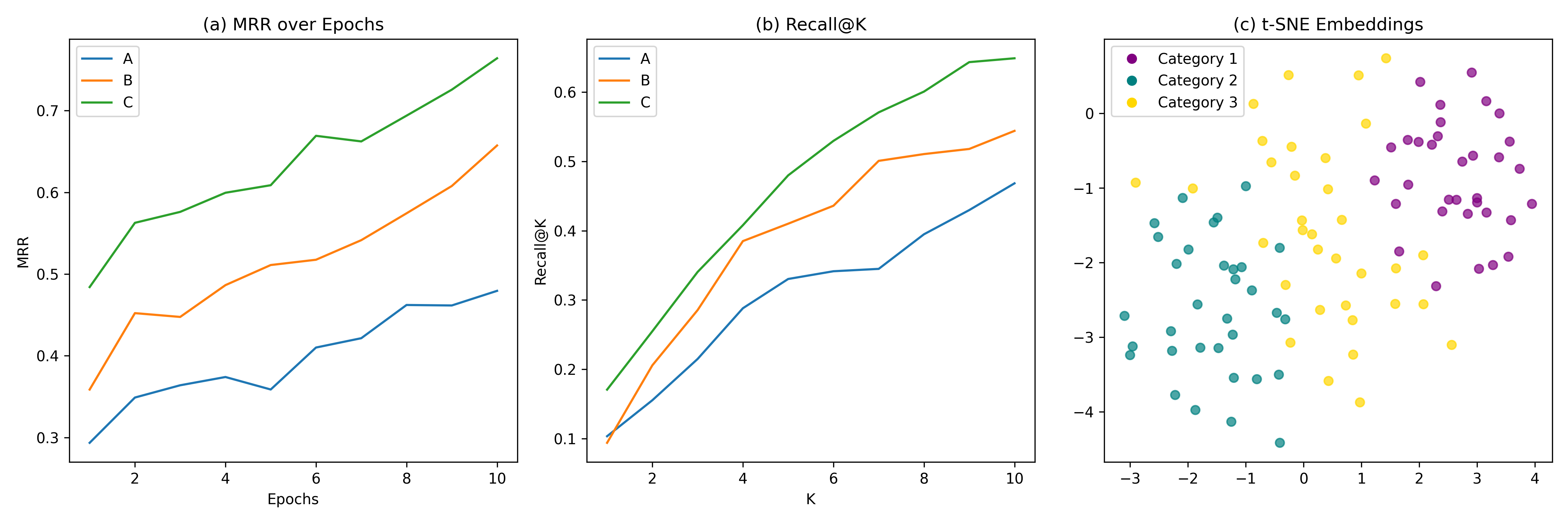}
	\caption{XMLC model evaluation. (a) MRR over training epochs. (b) Recall@K on the synthetic benchmark. (c) t-SNE projection of Model C embeddings, colored by ESCO Level-2 categories.}
	\label{fig:xmlc_fig}
\end{figure}

We next evaluate the downstream XMLC component, comparing Models A, B, and C on the synthetic hold-out benchmark. Figure~\ref{fig:xmlc_fig}(a) shows that Model C converges to the best MRR, followed by Model B and Model A. Figure~\ref{fig:xmlc_fig}(b) reports Recall@K, where Model C consistently attains higher recall across $K$, indicating more effective retrieval of relevant skills when expanding the candidate set.

Finally, Figure~\ref{fig:xmlc_fig}(c) visualizes a t-SNE projection of the learned embeddings (Model C), colored by ESCO Level-2 categories. While overlaps remain (as expected given semantically adjacent skill groups), the projection suggests meaningful category-level structure rather than degenerate separation.

\begin{table}[h]
\centering
\caption{Comparison of XMLC models on the synthetic validation set. Reported metrics align with Figure~\ref{fig:xmlc_fig}: MRR (final epoch) and Recall@K.}
\begin{tabular}{lccc}
\hline
\textbf{Model} & \textbf{MRR} & \textbf{Recall@5} & \textbf{Recall@10} \\
\hline
Model A (Standard, Single) & 0.48 & 0.33 & 0.47 \\
Model B (Optimized, Single) & 0.66 & 0.42 & 0.55 \\
\textbf{Model C (Optimized, Multi)} & \textbf{0.76} & \textbf{0.48} & \textbf{0.65} \\
\hline
\end{tabular}
\label{tab:xmlc_comp}
\end{table}

\subsection{Ablation Study}

\begin{figure}[h]
	\centering
	\includegraphics[width=\linewidth]{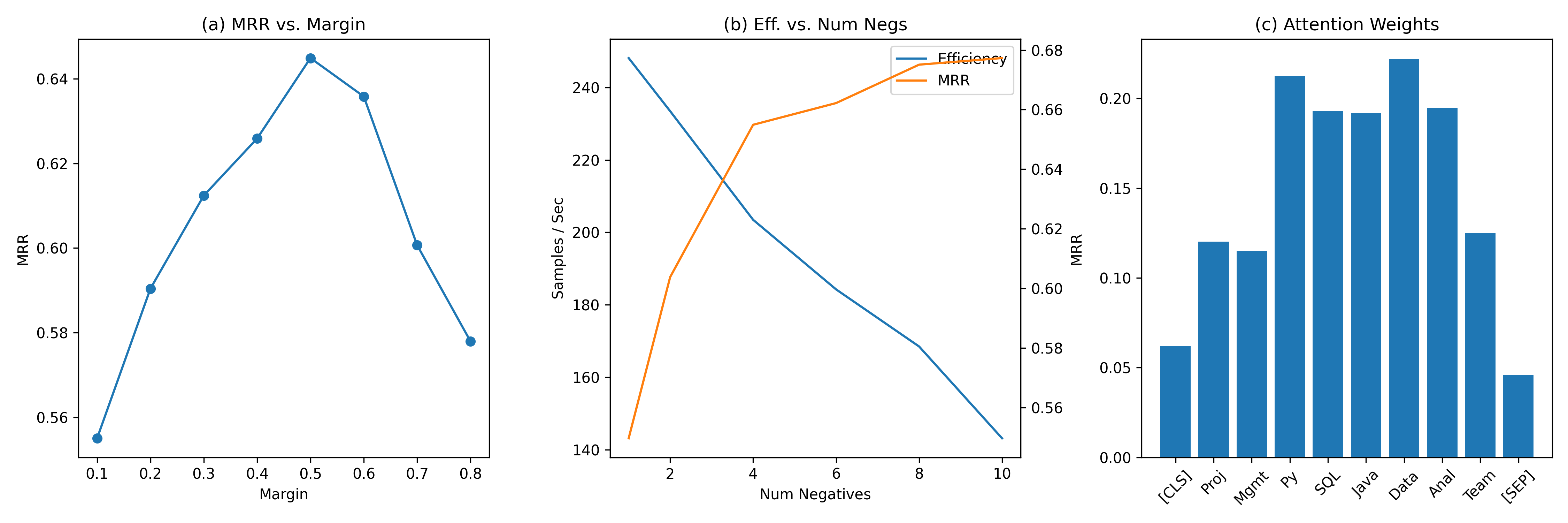}
	\caption{Ablation analysis. (a) MRR vs. contrastive margin. (b) Trade-off between training efficiency and MRR as the number of negative samples increases. (c) Relative attention scores over tokens, highlighting emphasis on skill-bearing terms.}
	\label{fig:ablation_fig}
\end{figure}

We conduct ablations to verify key design choices and characterize the sensitivity to contrastive-learning hyperparameters. Table~\ref{tab:ablation} and Figure~\ref{fig:ablation_fig}(a) show that performance is sensitive to the margin: MRR improves as the margin increases up to approximately $m=0.5$, after which overly aggressive separation degrades performance.

Figure~\ref{fig:ablation_fig}(b) shows diminishing returns as the number of negatives increases. While larger negative sets can improve MRR, training throughput declines substantially. In our setting, $N=8$ provides a practical operating point: it is near the MRR saturation region while avoiding the steepest efficiency loss at $N=10$.

Figure~\ref{fig:ablation_fig}(c) reports relative attention scores over tokens, indicating that the model assigns higher mass to skill-bearing terms (e.g., programming languages and task keywords) rather than delimiters.

\begin{table}[h]
\centering
\caption{Ablation analysis on Model B/C training choices. Throughput values correspond to training efficiency under the same hardware and batch configuration.}
\begin{tabular}{lcc}
\hline
\textbf{Configuration} & \textbf{MRR} & \textbf{Throughput (samples/s)} \\
\hline
No BiLSTM/Attention & 0.55 & -- \\
Margin = 0.1 & 0.55 & -- \\
Margin = 0.5 & 0.65 & -- \\
Margin = 0.8 & 0.58 & -- \\
Num\_Negatives = 1 & 0.55 & 250 \\
Num\_Negatives = 8 & 0.68 & 170 \\
Num\_Negatives = 10 & 0.68 & 150 \\
\textbf{Selected (Margin = 0.5, Neg = 8)} & \textbf{0.65} & \textbf{170} \\
\hline
\end{tabular}
\label{tab:ablation}
\end{table}

\subsection{End-to-End Zero-Shot Transfer (Real Data)}

\begin{figure}[h]
	\centering
	\includegraphics[width=\linewidth]{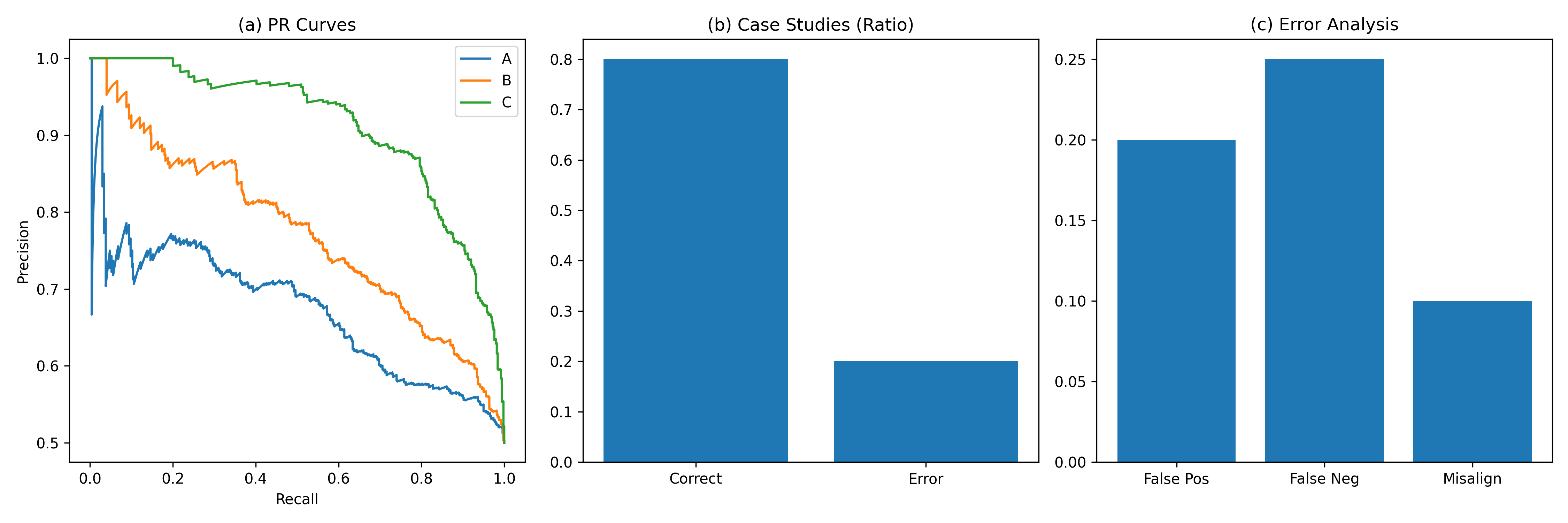}
	\caption{End-to-end transfer performance on real-world test data. (a) Precision--Recall curves for Models A/B/C. (b) Correct vs. error ratio in qualitative case studies. (c) Error breakdown into false positives, false negatives, and taxonomy misalignment.}
	\label{fig:end_to_end_fig}
\end{figure}

We evaluate end-to-end zero-shot transfer on manually annotated job advertisements. Figure~\ref{fig:end_to_end_fig}(a) shows that Model C dominates Models A and B in Precision--Recall space, indicating stronger zero-shot generalization from synthetic supervision to real-world text. Qualitative case inspection (Figure~\ref{fig:end_to_end_fig}(b)) suggests that the majority of predictions are correct, while the remaining errors concentrate in a smaller fraction of challenging samples.

Error analysis in Figure~\ref{fig:end_to_end_fig}(c) indicates that false negatives are the most frequent failure mode, followed by false positives. ``Misalignment'' errors (cases where the model retrieves a closely related but not exactly matching skill label) are less frequent but remain a substantive source of residual error, consistent with the inherent granularity and overlap of skill taxonomies.

\begin{table}[h]
\centering
\caption{End-to-end zero-shot performance on real-world job ads. We report AUPRC (Average Precision) and F1 at the default threshold for comparability with Figure~\ref{fig:end_to_end_fig}(a).}
\begin{tabular}{lcc}
\hline
\textbf{Model} & \textbf{AUPRC} & \textbf{F1} \\
\hline
Model A & 0.72 & 0.69 \\
Model B & 0.82 & 0.75 \\
\textbf{Model C} & \textbf{0.90} & \textbf{0.80} \\
\hline
\end{tabular}
\label{tab:end_to_end}
\end{table}

\subsection{Scalability and Robustness}

\begin{figure}[h]
	\centering
	\includegraphics[width=\linewidth]{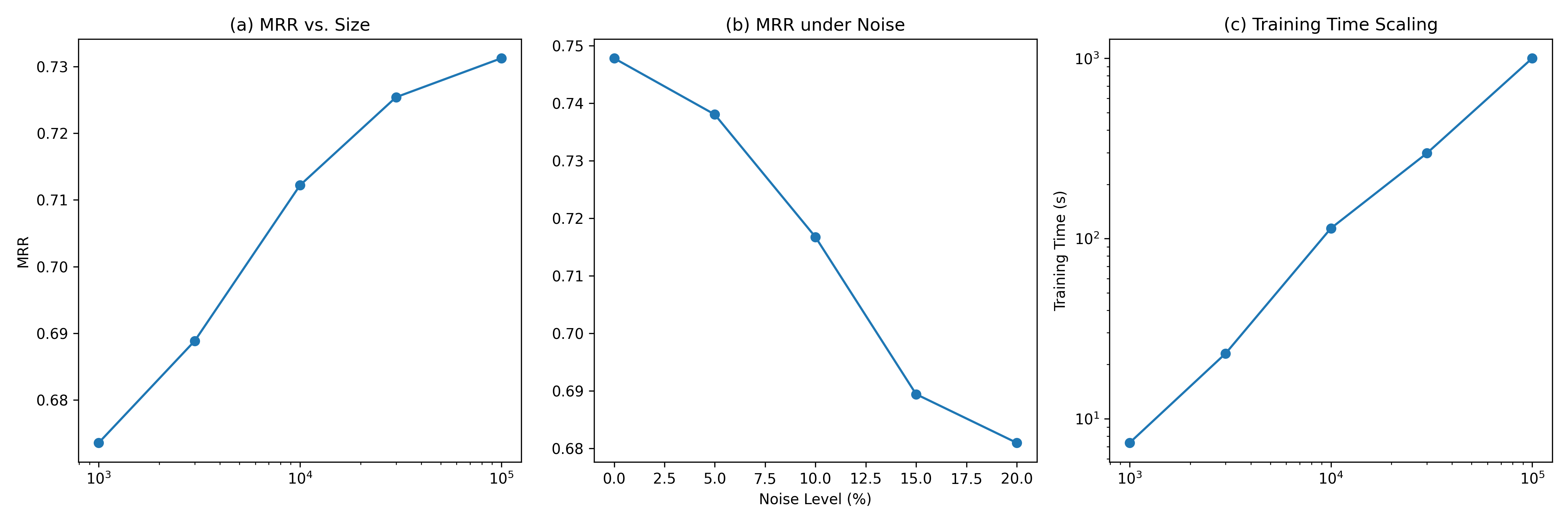}
	\caption{Scalability and robustness analysis. (a) MRR vs. synthetic training set size. (b) MRR under increasing noise levels (0--20\%). (c) Training time scaling with dataset size (log--log).}
	\label{fig:scalability_fig}
\end{figure}

Finally, we evaluate scalability and robustness for Model C. Figure~\ref{fig:scalability_fig}(a) shows that MRR improves with synthetic dataset size, with gains becoming smaller at larger scales (diminishing marginal returns). Figure~\ref{fig:scalability_fig}(b) shows graceful degradation under increasing noise, indicating that the model is not overly brittle to moderate text corruption. Figure~\ref{fig:scalability_fig}(c) demonstrates approximately linear scaling of training time with dataset size on a log--log scale, consistent with expectations for minibatch training under fixed hyperparameters.

\section{Conclusion} \label{sec:conclusion}

This paper targets the practical challenge of mapping unstructured job-ad text to a standardized skills ontology (ESCO) in settings where manually annotated training data are scarce or unavailable. We propose a zero-shot XMLC framework that couples LLM-based synthetic supervision with contrastive learning for retrieval-style skill prediction. A key design choice is \emph{hierarchically constrained} generation: by sampling multi-skill sentences under ESCO Level-2 constraints, we obtain synthetic training signals that are more coherent and more discriminative than unconstrained alternatives.

Across a suite of intrinsic and extrinsic evaluations, the results support three main conclusions. First, Level-2 constrained generation produces synthetic text with improved fluency (lower perplexity) and stronger separability in downstream diagnostics, suggesting reduced noise in the resulting supervision. Second, the RoBERTa-based sentence filter provides an effective front-end for excluding non-skill content, achieving balanced false positive/false negative behavior and a consistently stronger precision--recall profile than keyword matching, at the cost of higher inference latency. Third, in the XMLC stage, the optimized multi-label model (Model~C) converges to the best ranking quality and recall-at-$K$ on the synthetic benchmark, and it also exhibits superior end-to-end precision--recall behavior on real job ads. The remaining errors are dominated by false negatives and taxonomy misalignment cases, highlighting the inherent ambiguity of fine-grained skill definitions and the long-tail nature of job-ad phrasing.

Overall, the proposed framework offers a scalable pathway for labor market intelligence: it enables automated projection of large volumes of job postings into standardized skill representations without requiring labeled training data. Future work will extend the pipeline in three directions: (i) multilingual transfer and cross-lingual alignment for non-English job markets, (ii) hierarchy-aware learning objectives (e.g., graph-based regularization or taxonomy-consistent decoding) to better exploit ESCO structure, and (iii) generalization to additional taxonomies (e.g., O*NET) and domain-specific skill ontologies.

\bibliographystyle{unsrtnat}
\bibliography{references}  

\appendix

\section{Synthetic Data Generation Details}

To augment our dataset with diverse and realistic sentence-skill pairings, we employed a large language model (LLM) for synthetic data generation. We utilized the DeepSeek V3.1 model (accessed via the `deepseek-chat` API endpoint), leveraging its strong multilingual capabilities to generate contextually relevant Chinese job advertisement sentences based on ESCO skill descriptions.

\subsection{LLM Configuration}
Based on the generation scripts, the model was configured with the following hyperparameters to ensure output diversity while preventing repetition:

\begin{itemize}
    \item \textbf{Model}: DeepSeek V3.1 Chat
    \item \textbf{Temperature}: 0.9 
    \item \textbf{Presence Penalty}: 0.8 
    \item \textbf{Prompting Strategy}: Few-Shot Learning (We provided specific examples of skill-to-sentence mapping within the conversation history to guide the model's style).
\end{itemize}

\subsection{Prompt Templates}

We utilized a role-playing strategy where the system was instructed to act as an "experienced recruitment ad copywriter." Below are the specific templates used for single-label, multi-label, and non-skill sentence generation.

\subsubsection{System Instruction}
All generation tasks shared the following system-level instruction:

\begin{quote}
\textit{System:} "You are an experienced recruitment ad copywriter expert, good at generating recruitment ad sentences based on skills. Please respond by generating sentences used in hypothetical recruitment ads based on user requirements, representing the demand for specific skills. Ensure diversity in the generated sentences and do not repeat sentences or structures."
\end{quote}

\subsubsection{Single-Label Generation}
For the single-label dataset (\texttt{esco\_sentence\_one.csv}), we provided the model with the ESCO skill label and definition. The prompt included few-shot examples (e.g., for "Project Management" and "Java") to demonstrate the desired output format.

\begin{verbatim}
User: 
Number of sentences: [N] 
Skill: [translated_skill] 
Definition: [description_translation]
\end{verbatim}

\subsubsection{Hierarchical Multi-Label Generation}
For the multi-label dataset (\texttt{esco\_sentence\_two.csv}), the prompt was designed to integrate two skills into a single context. As detailed in the methodology, these pairs were pre-selected to belong to the same ESCO Level-2 category.

\begin{verbatim}
User: 
Number of sentences: [N] 
Skill 1: [skill_1] 
Definition 1: [description_1] 
Skill 2: [skill_2] 
Definition 2: [description_2]
\end{verbatim}

\subsubsection{Negative (No-Skill) Sample Generation}
To train the binary filter, we also generated sentences typical of job ads that do not contain skill requirements (e.g., salary, benefits, company culture).

\begin{verbatim}
User: 
Number of sentences: [N] 
Do not include any requirements regarding labor skills.
\end{verbatim}

\subsection{Data Post-Processing}
The raw outputs were parsed to remove bullet points (e.g., "- ") and filtered to remove duplicates. We generated approximately 100 sentences per prompt for the single-label task and smaller batches for the multi-label task, resulting in the final datasets described in Section \ref{sec:method}.





\end{document}